\documentclass[runningheads]{llncs}
\usepackage{amssymb}
\usepackage{amsmath}
\usepackage{graphicx}
\usepackage{url}
\usepackage[export]{adjustbox}
\usepackage{array}
\usepackage[labelformat=empty]{subfig}
\usepackage{color,soul} 
\soulregister\ref{7} 
\soulregister\cite{7} 
\usepackage{float} 
\usepackage{hyperref}
\hypersetup{colorlinks}
\usepackage[colorinlistoftodos]{todonotes}
\usepackage{booktabs} 
\usepackage{rotating}
\usepackage{multirow}
\usepackage{transparent}
\usepackage{tablefootnote}
\usepackage{algorithm}
\usepackage{algpseudocode}
\usepackage{todonotes}

\usepackage[subtle,margins=normal,leading=normal]{savetrees} 
\usepackage{bm} 
\usepackage{colortbl}
\usepackage{tabulary}

\begin{document}
%

\title{Split-U-Net: Preventing Data Leakage in Split Learning for Collaborative Multi-Modal Brain Tumor Segmentation}
\titlerunning{Split-U-Net}

\titlerunning{Split-U-Net}
%
\author{Holger R. Roth, Ali Hatamizadeh, Ziyue Xu, \\
Can Zhao, Wenqi Li, Andriy Myronenko, Daguang Xu}
\authorrunning{H. Roth et al.}
%
\institute{NVIDIA, Bethesda, USA}
\maketitle              
%
\begin{abstract}
Split learning (SL) has been proposed to train deep learning models in a decentralized manner. For decentralized healthcare applications with vertical data partitioning, SL can be beneficial as it allows institutes with complementary features or images for a shared set of patients to jointly develop more robust and generalizable models. In this work, we propose ``Split-U-Net'' and successfully apply SL for collaborative biomedical image segmentation. Nonetheless, SL requires the exchanging of intermediate activation maps and gradients to allow training models across different feature spaces, which might leak data and raise privacy concerns. Therefore, we also quantify the amount of data leakage in common SL scenarios for biomedical image segmentation and provide ways to counteract such leakage by applying appropriate defense strategies.
\end{abstract}
%
%

\section{Introduction}
\label{sec:intro}
Collaborative and decentralized techniques to train artificial intelligence (AI) models have been gaining popularity, especially in healthcare applications where data sharing to build centralized datasets is particularly challenging due to patient privacy and regulatory concerns~\cite{rieke2020future}. Federated learning (FL)~\cite{mcmahan2017communication} and split learning (SL)~\cite{gupta2018distributed} are two approaches that can be useful depending on the nature of the data partitioning~\cite{yang2019federated}. 
In the healthcare and biomedical imaging sector, data is often ``horizontally'' partitioned such that each participating site, i.e., a hospital, possesses some data/features and optionally corresponding labels for their set of patients. Horizontal FL (HFL) algorithms like federated averaging~\cite{mcmahan2017communication} typically train models initialized from a current ``global'' model independently on each participant and frequently update the global model with the model gradients sent by each site. 
In contrast, so-called ``vertical'' data partitioning allows sites with complementary features but from an overlapping set of patients to collaborate~\cite{yang2019federated}. This vertical FL (VFL) scenario could be useful where different sites possess features that need to be securely combined in order to train a joined AI model, e.g., one hospital has imaging while the other one has lab results or the diagnoses for the same set of patients. Here, SL can be used to train models when using deep learning (DL) methods for VFL. During training, SL splits the forward pass of a DL model into two or more parts and exchanges intermediate features or activation maps and gradients between participating sites to complete a training step. Therefore features from different sites can be combined in later parts of the network and the model can be trained across institutional boarders~\cite{vepakomma2018split}. 

\begin{figure}[htbp]
    \centering
    \includegraphics[width=0.6\columnwidth]{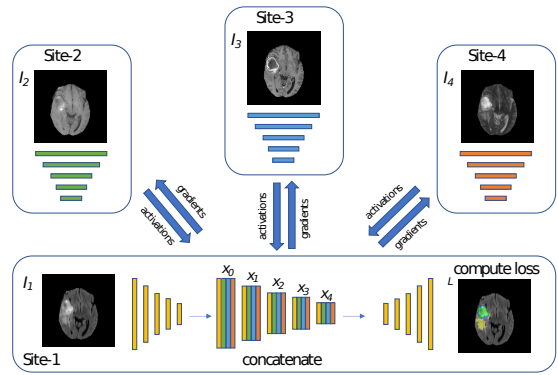}
    \caption{Split learning set up with Split-U-Net. \label{fig:setup}}
\end{figure}

In biomedical image segmentation, VFL could be useful to combine different image modalities of the same patient in order to train joined segmentation models collaboratively. This scenario is what we explore in this work by studying SL as a collaborative technique to learn a tumor segmentation model for multi-model MRI images. For this purpose, we propose ``Split-U-Net'', a modification to the popular U-Net~\cite{ronneberger2015u} architecture, to allow its use in a VFL setup. Figure ~\ref{fig:setup} illustrates the situation where four sites would like to jointly train a multi-modal segmentation model given their corresponding images. Only one site possesses the label mask and computes the loss to be optimized. 
Previous works on SL in healthcare applications have focused on classification and regression tasks~\cite{vepakomma2018split,poirot2019split,ha2022feasibility}.
One example of splitting U-Net for single modality semantic segmentation was described in~\cite{poirot2020split}, but to the best of our knowledge, our work is the first to apply SL in a multi-modal vertical data partitioning scenario for biomedical image segmentation across multiple parties.

We show that SL can be used successfully for this task and also investigate potential security implications that arise from sharing intermediate features between collaborating sites by implementing an effective inversion attack. Prior works on inversion attacks in SL were mainly focused on images of small sizes (MNIST or CIFAR-10)~\cite{pasquini2021unleashing,erdogan2021unsplit,he2019model,jin2021catastrophic} and are theoretical in nature. However, it is important for the medical imaging community to understand the potential benefits and security considerations for applying SL in healthcare applications. Our inversion attack not only shows the potential risks but can be used to quantify and inform appropriate defense strategies against it. In this work, we explore both dropout~\cite{srivastava2014dropout} and differential privacy (DP)~\cite{dwork2006calibrating} to prevent data leakage during SL. 
Our contributions can be summarized as follows.
\begin{itemize}
    \item We propose ``Split-U-Net'' and successfully apply SL for biomedical image segmentation for multi-institutional collaboration.
    \item We develop a successful inversion attack to measure and quantify data leakage in SL.
    \item We propose and evaluate defense measures (dropout and DP) to prevent data leakage.
\end{itemize}

\section{Methods}
\label{sec:methods}
\newcommand\fenc{f}
\newcommand\fencattack{\tilde{f}}
\newcommand\fdec{g}
\subsection{Split-U-Net}
The basis of our network is a common implementation of U-Net~\cite{ronneberger2015u} with $L=4$ down- and up-sampling levels. The default number of output features at each level are configured as shown in Table~\ref{tab:network}. To turn this network $F(x)$ into Split-U-Net $F(x)=\fdec(\fenc(x))$ used for multi-modal collaborative SL, we divided the number of encoder features by the number of sites/modalities $K$ involved in training. The number of features in the decoder stays the same as in the default network. Fig.~\ref{fig:setup} shows an example setup with $K=4$. In this example, the ``split'' is done at the bottleneck. All encoder layers participate in SL, and only the site with label images has the decoder for segmentation.
%
\begin{table}[b]
\caption{U-Net and Split-U-Net features for brain tumor segmentation. \label{tab:network}}
\scriptsize
\centering
    \begin{tabular}{|l|l|rrrrr|rrrr|r|}
    \toprule
    Level $i$                                   &In & 0 & 1 & 2 & 3 & 4 & 5 & 6 & 7 & 8 & Out \\
    \hline
    U-Net (default) $F(x)$                            & 4 & 32& 32& 64& 128& 256& 128& 64& 32& 32 & 4 \\
    Split-U-Net Encoder (per site) $\fenc^k(x)$ & 1 &  8&  8& 16&  32&  64&   -&  -&  -&   -& - \\
    Split-U-Net Decoder $\fdec(x)$              & - &  -&  -&  -&   -&   -& 128& 64& 32& 32 & 4 \\
    \bottomrule
    \end{tabular}
\end{table}
During training, each site $k$ computes the forward pass $\left\{x_0^k, x_1^k, \dots, x_{L}^k\right\} = \fenc^k(I_k)$ where $I_k$ is a mini-batch of size $B$ of input images with modality $k$ and $x_i^k$ is the feature map of layer $i$ of $L$. Corresponding batch indices are communicated to each site before each training step. The site $k$ with label images then takes the activation maps from all other participating sites and concatenates them at the appropriate feature levels (see Fig.~\ref{fig:setup}). It then computes the loss and backward pass to obtain a gradient
\begin{equation}
\scriptsize
\nabla \leftarrow \mathcal{L}_{seg}\left(\fdec(\left\{x_0^k, x_1^k, \dots, x_{L}^k\right\}), labels\right)
\end{equation}
 and performs an optimizer update on its part of the network $\fdec(x)$.
The gradient $\nabla$ at the split level is then communicated back to all $\fenc^k(x)$ to complete the backward pass and update their parts of the model (the encoder branches $\fenc^k(x)$), and the process is iterated until convergence. Note that in SL, a larger batch size can be used to reduce the total number of communication steps needed~\cite{singh2019detailed}.
We assume that at each iteration, a random set of batch indices $b_i$ is selected such that each client uses the same patients' data and augmentation to build their mini-batch $I_k$. Furthermore, each site might apply additional spatial normalization steps, e.g., using non-linear image registration to bring their images from different modalities into a common data space to help the network better encode common anatomical features across modalities~\cite{studholme1997automated,xu2015efficient}.
\subsection{Measuring data leakage by inversion attack}
In SL, activation maps are shared to complete each iteration step~\cite{gupta2018distributed}. Therefore, a potential malicious actor, e.g., Site-1 in Fig. ~\ref{fig:setup}, receiving the activation maps from other sites may invert them to recover the underlying private data. Such attacks used to recover the data are called ``inversion attacks''~\cite{usynin2021adversarial}. In the following, we explain how our inversion attack is executed and how its result can be used to measure the data leakage in order to inform an appropriate defense strategy.
Our attack tries to optimize a randomly initialized $C$-channel input $\tilde{I}_i$ such that the activations $\tilde{x_i}$ at the forward layer of the attacker model $\fencattack_i(x)$ become the same as the intercepted activations ${x_i}$. In this work, we assume the attacker has access to the current state of the model used by the client to generate the forward pass. Therefore $\fencattack_i^k(x) \equiv \fenc_i^k(x)$ given the same input $x$. This setting is typically referred to as a ``white-box'' attack~\cite{he2019model}. In practical implementations of SL, this could be the case if a common network is used to initialize $\fenc^k(x)$ on each participating client. The main loss used to align both activation maps is a $L^2$-norm. Furthermore, we employ two common image prior losses often used in inversion attacks~\cite{geiping2020inverting,yin2021see}, namely total variation~\cite{rudin1992nonlinear} (TV) and $L^2$-norm of the recovered image $\tilde{I}$. The main loss for the inversion attack hence becomes
\begin{equation}
\scriptsize
  \mathcal{L_\text{inv}}\left(x_i^k, \tilde{x}_i^k, \tilde{I}_i^k\right) \ = \ \alpha_{\text{act}} || x_i^k - \tilde{x}_i^k ||_2 \ + \ \alpha_{\text{tv}} TV(\tilde{I}_i^k) \ + \ \alpha_{l_2} || \tilde{I}_i^k ||_2.
  \label{equ:inv_ioss}
\end{equation}
Therefore, the final inversion attack to recover an image $\tilde{I}_i$ from activation $x_i$ at level $i$ of Split-U-Net can be formulated as
\begin{equation}
\scriptsize
  \tilde{I}_i = \underset{\tilde{I}}{\mathrm{argmin}} \ \mathcal{L_\text{inv}}\left(x_i^k, \tilde{x}_i^k, \tilde{I}_i^k\right),
  \label{equ:inv_img}
\end{equation}
where $\tilde{I}_i \in \mathbf{R}^{B, C, H, W}$ with $B, C, H, W$ being the batch size, number of channels, height and width of the image, respectively. 
Note that the inversion can be run on large batch sizes $B$ or independently for each activation in a mini-batch, depending on the compute resources of the attacker. The data inversions from intercepted activation maps can be seen in Fig.~\ref{fig:skip_invs}.
To measure the amount of data leakage, we compute a common similarity metric between the recovered image $\tilde{I}_i$ and the original image $I$ used to produce the activation $x_i$. Structural Similarity index (SSIM)~\cite{wang2004image} aims to provide a more intuitive and interpretable metric compared to other commonly used metrics like root-mean-squared error or peak signal-to-noise ratio. We, therefore, use SSIM in our analysis, but including other metrics would be possible.
\subsection{Defenses}
\label{sec:defenses}
\begin{figure}[htbp]
\begin{minipage}{.48\textwidth}
\centering
\scriptsize
  (a) originals [1, 192, 192]\hfill \\
  \includegraphics[width=1.0\linewidth]{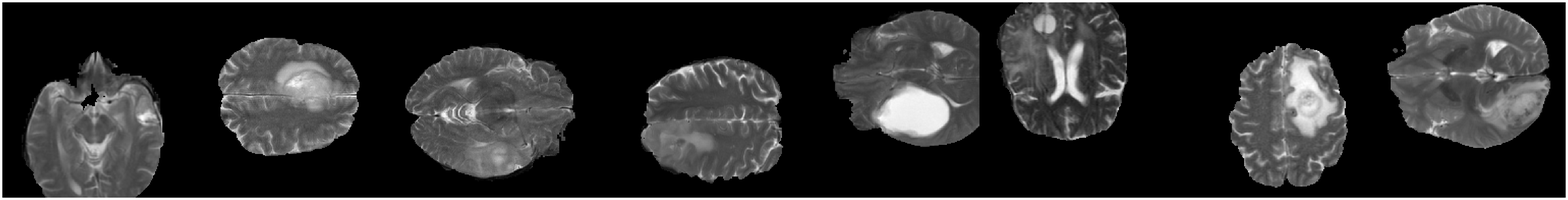} \\
  (b) activations $x_0$ [8, 192, 192]\hfill \\
  \includegraphics[width=1.0\linewidth]{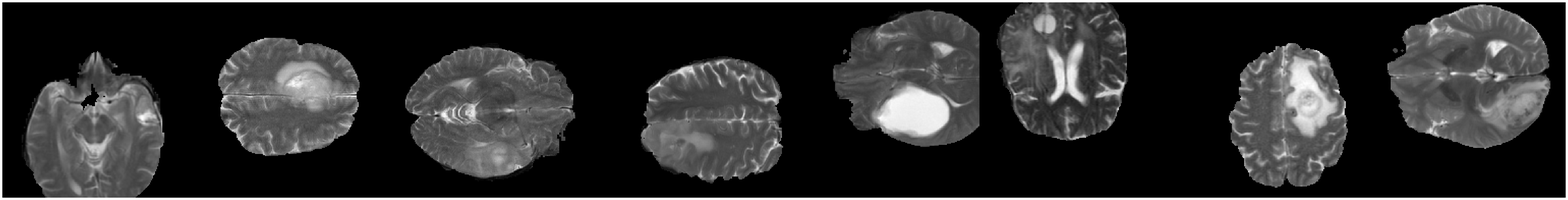} \\
  (c) activations $x_1$ [8, 96, 96]\hfill \\
  \includegraphics[width=1.0\linewidth]{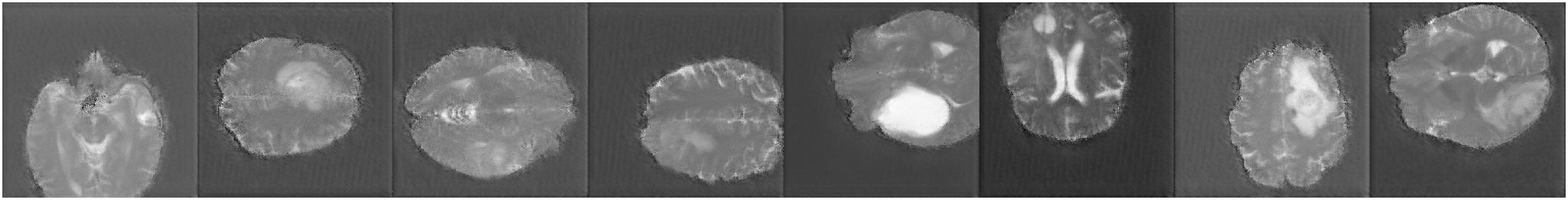} \\
\end{minipage}\hfill
\begin{minipage}{.48\textwidth}
\centering
\scriptsize
  (e) activations $x_2$ [16, 48, 48]\hfill \\
  \includegraphics[width=1.0\linewidth]{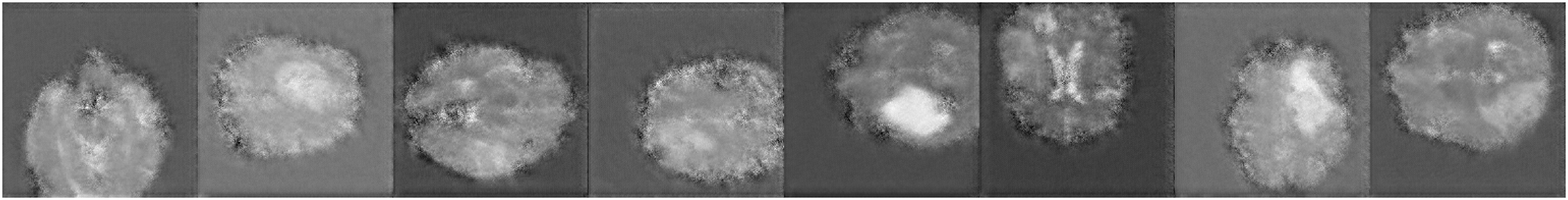} \\
  (f) activations $x_3$ [32, 24, 24]\hfill \\
  \includegraphics[width=1.0\linewidth]{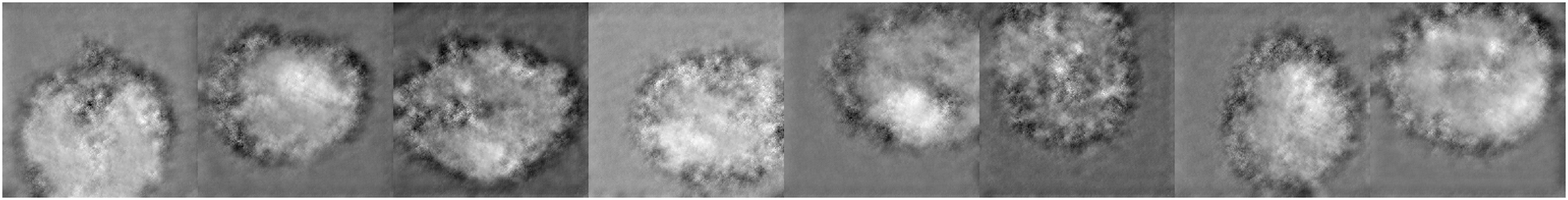} \\
  (g) activations $x_4$ [64, 12, 12]\hfill \\
  \includegraphics[width=1.0\linewidth]{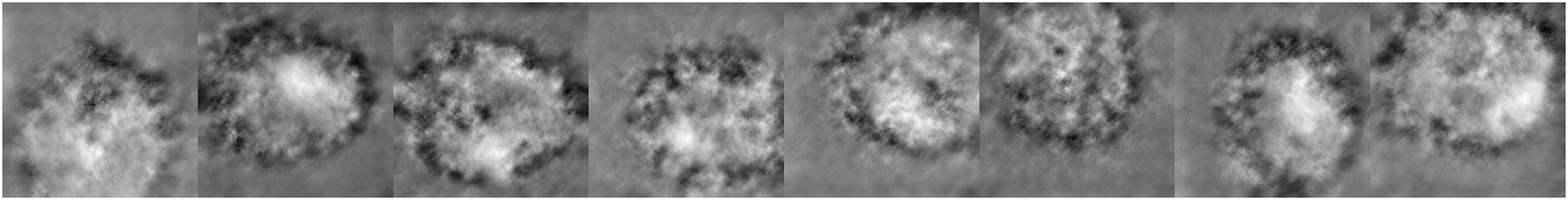}
\end{minipage}  
\caption{Inversions from activations sent from different layers of the Split-U-Net encoder of Site-4 possessing one MRI modality when training with a mini-batch size of 8. Activations from earlier layers from the encoder are more likely to leak data, i.e., $x_0 \sim x_2$. The inversions from other sites and modalities are of the same quality. \label{fig:skip_invs}}
\end{figure}

A straightforward defense strategy is to not send feature activation maps from early layers ($x_0$, $x_1$, and $x_2$) which are likely to leak more data (see Fig.~\ref{fig:skip_invs} and Fig.~\ref{fig:ssim}).
We also investigate dropout~\cite{srivastava2014dropout} as an effective tool against inversion attacks. Each layer of the encoder can randomly drop the activations from neurons of the network with a probability of $p_\mathrm{dropout}$. 
Another effective tool often used in the FL literature~\cite{li2019privacy,kaissis2021end,hatamizadeh2022gradient,yang2020local}, is differential privacy (DP). DP in its simplest form adds some calibrated random noise to any shared values in order to preserve the privacy of individual data entries. Here, we use a Gaussian mechanism~\cite{yang2020local} to add random noise sampled from a normal distribution $N\left(0,\ \sigma^2 \right)$ to each activation mask $x_i^k$ before sharing it with the next participant.

\section{Experiments \& Results}
\label{sec:results}
\begin{figure}[htbp]
\centering
\newcommand{\trimr}{0cm} 
\begin{minipage}{.48\textwidth}
    \centering
    \scriptsize
    (a) activations $x_0$ ($p_{\mathrm{dropout}}=0.1$) \\
    \includegraphics[trim={0 0 {\trimr} 0},clip,width=.95\linewidth]{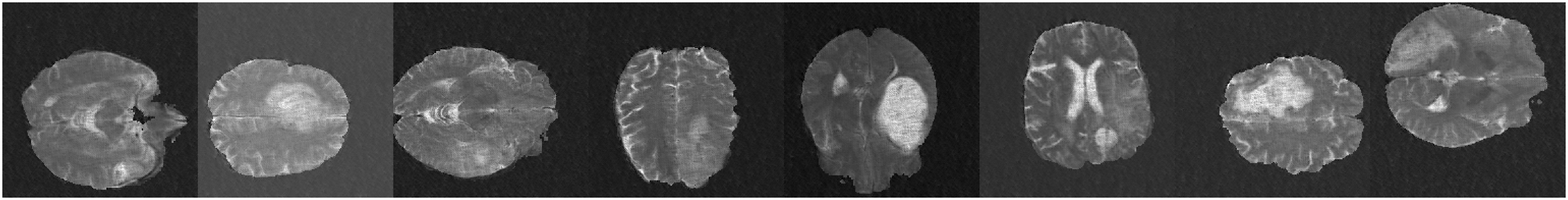} \\
    (b) activations $x_0$ ($p_{\mathrm{dropout}}=0.2$) \\
    \includegraphics[trim={0 0 {\trimr} 0},clip,width=.95\linewidth]{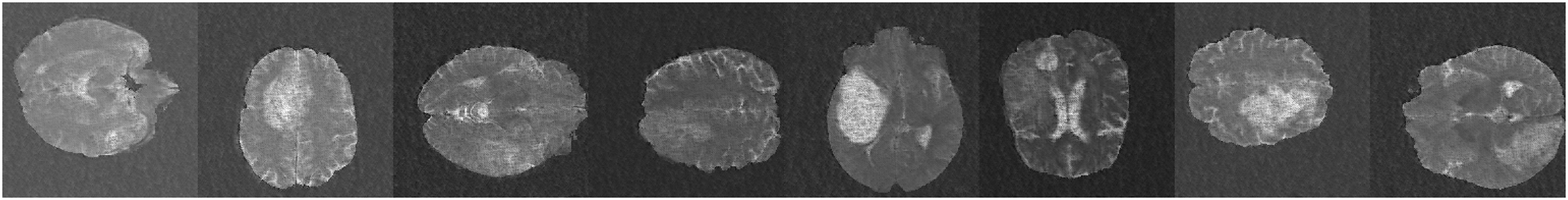} \\
    (c) activations $x_0$ ($p_{\mathrm{dropout}}=0.5$) \\
    \includegraphics[trim={0 0 {\trimr} 0},clip,width=.95\linewidth]{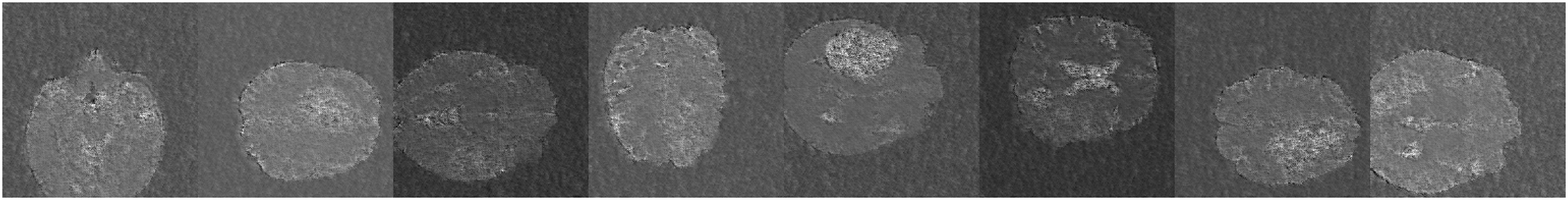} \\  
    (d) activations $x_0$ ($p_{\mathrm{dropout}}=0.8$) \\
    \includegraphics[trim={0 0 {\trimr} 0},clip,width=.95\linewidth]{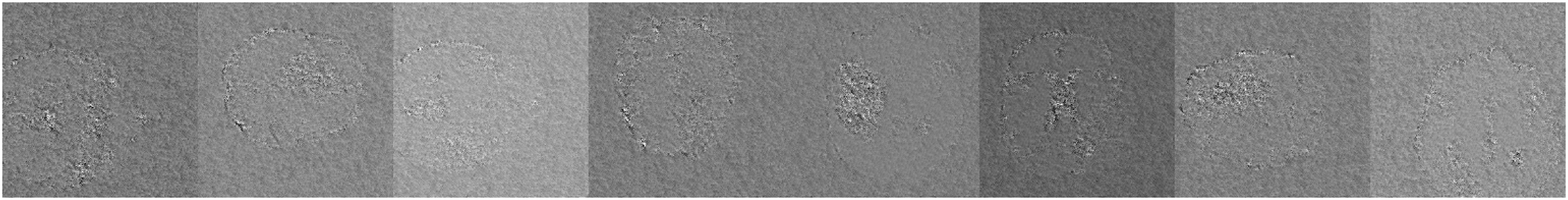}
    (e) activations $x_0$ ($\sigma=1$) \\
    \includegraphics[trim={0 0 {\trimr} 0},clip,width=.95\linewidth]{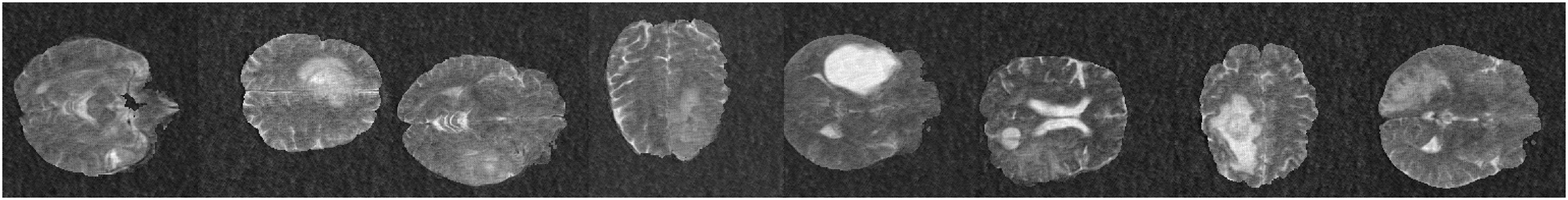} \\
    (f) activations $x_0$ ($\sigma=2$) \\
    \includegraphics[trim={0 0 {\trimr} 0},clip,width=.95\linewidth]{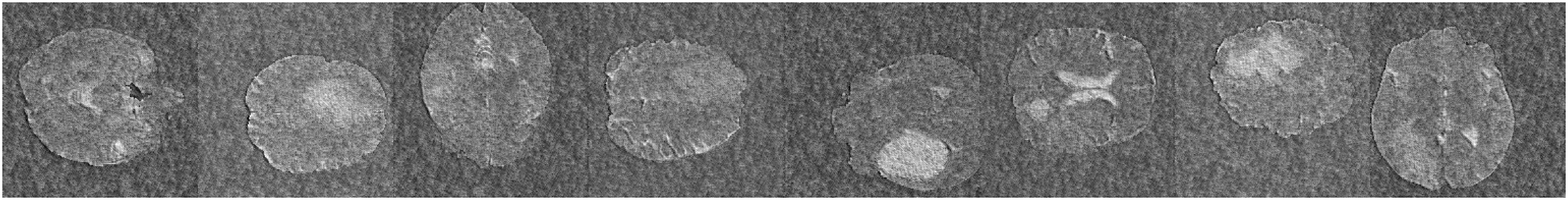} \\
    (g) activations $x_0$ ($\sigma=3$) \\
    \includegraphics[trim={0 0 {\trimr} 0},clip,width=.95\linewidth]{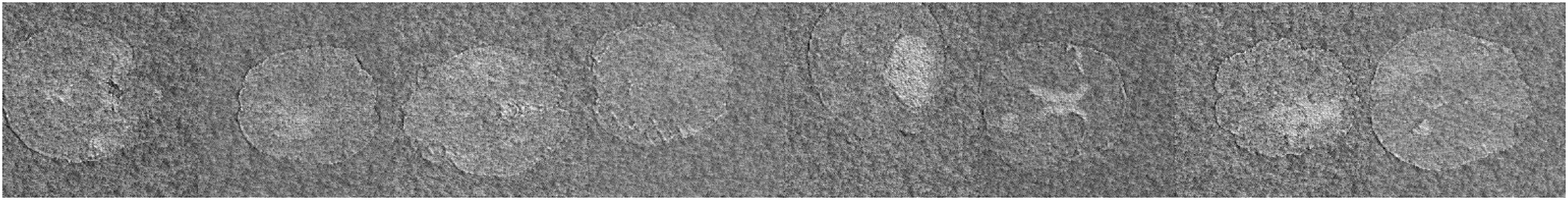} \\
    (h) activations $x_0$ ($\sigma=5$) \\
    \includegraphics[trim={0 0 {\trimr} 0},clip,width=.95\linewidth]{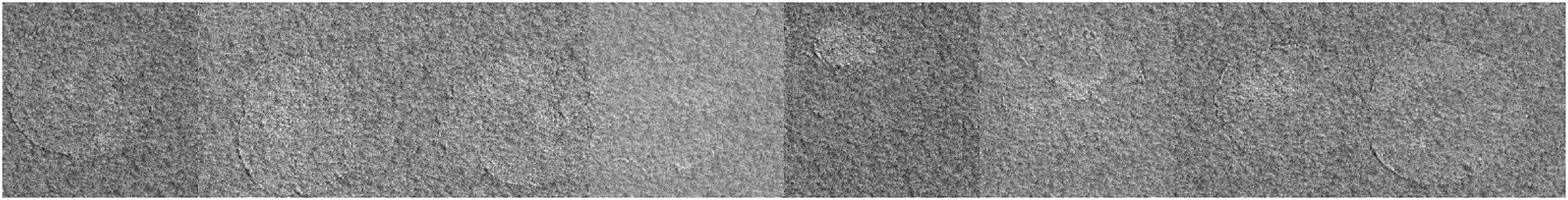}
    \caption{Dropout (a-d) and differential privacy (e-h) as a defense against inversion attacks. \label{fig:defense_inv}}
\end{minipage}\hfill
\begin{minipage}{.48\textwidth}
    \centering
    \scriptsize
    \centering    
    \includegraphics[width=.95\linewidth]{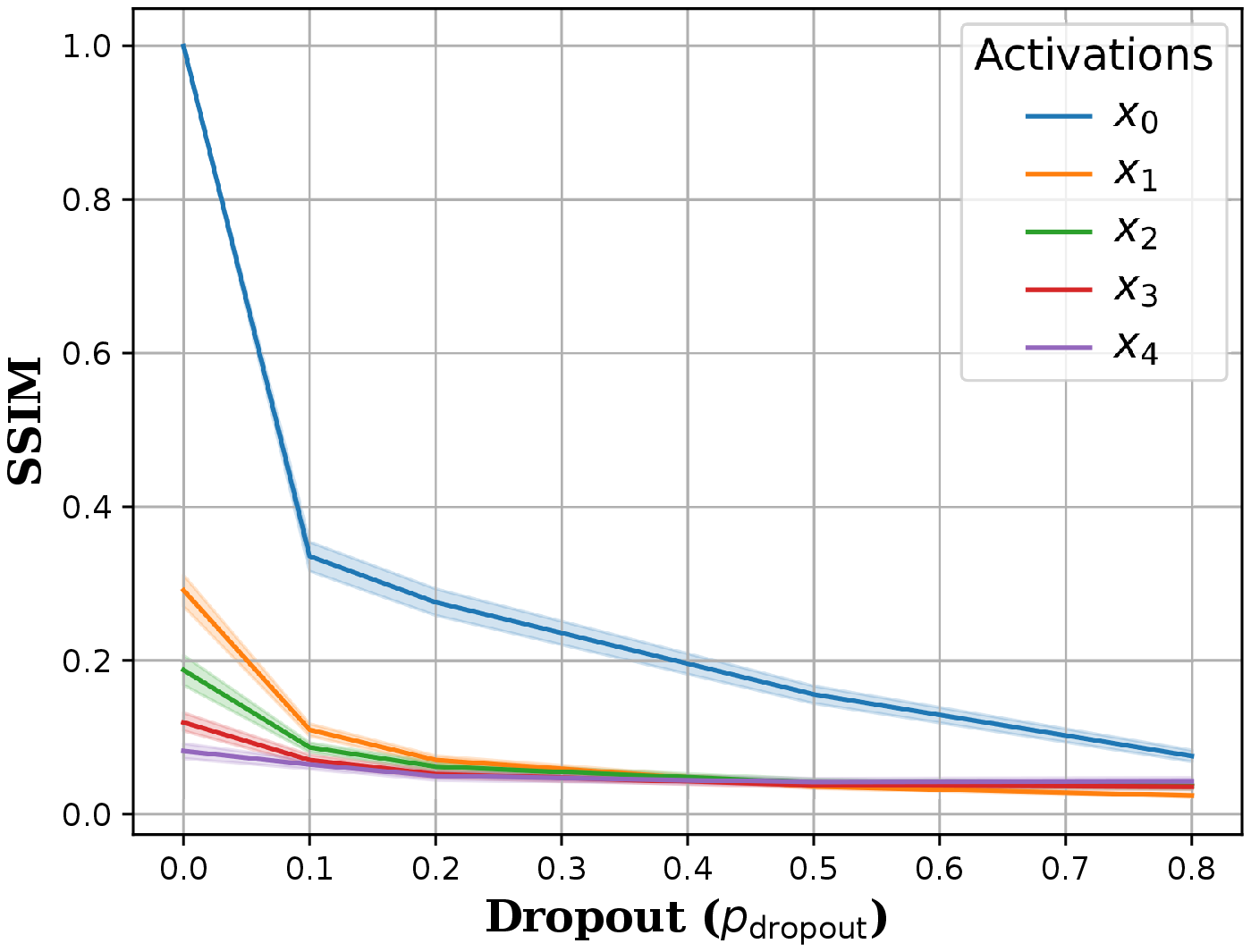}
    \includegraphics[width=.95\linewidth]{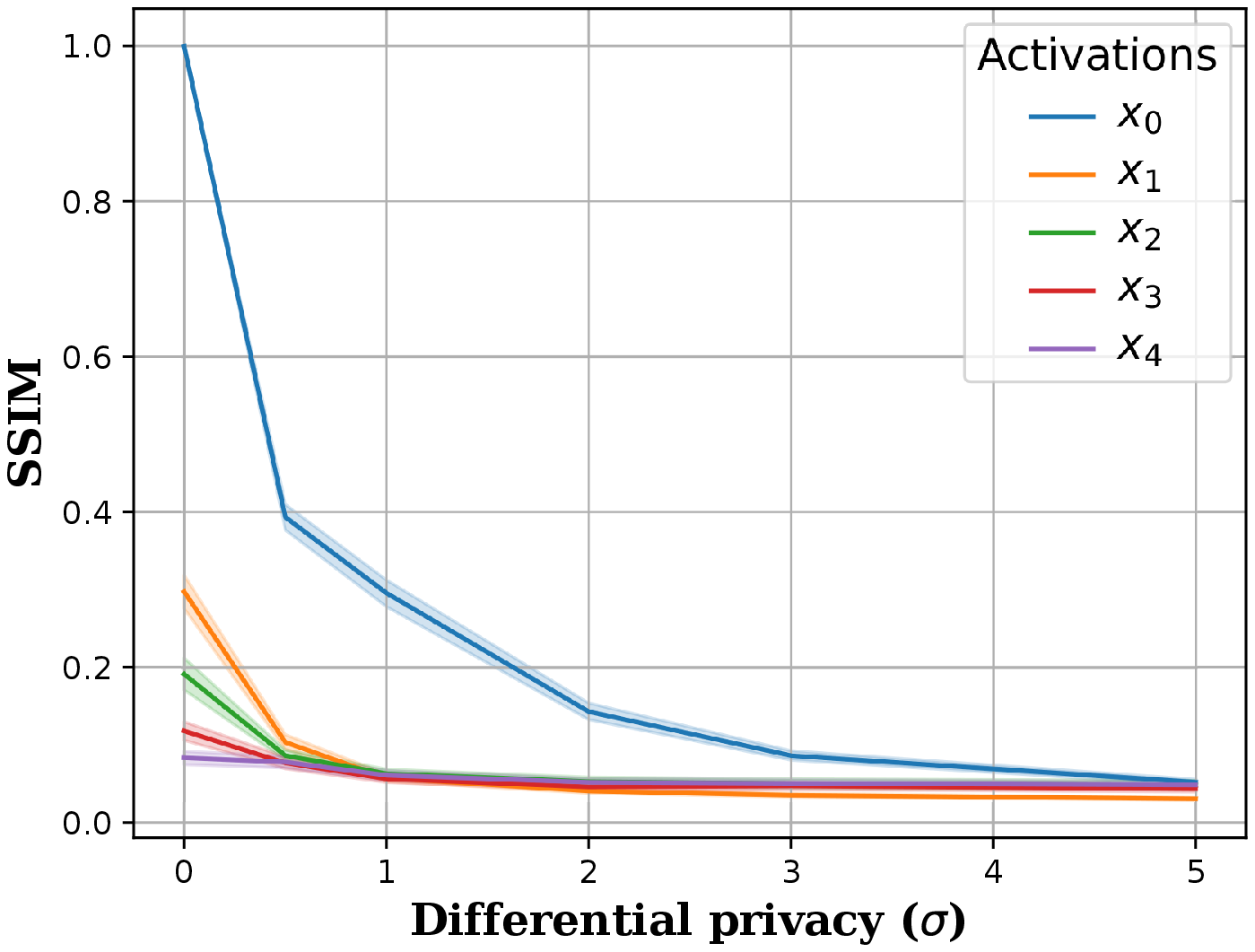} 
    \caption{Structural SIMilarity index (SSIM)~\cite{wang2004image} between the original images and inversions of each activation.\label{fig:ssim}}
\end{minipage}
\end{figure}

\paragraph{\textbf{Data:}}
In our study, we assume a collaborative model training setup where four institutes, here referred to as ``sites'', jointly train a multi-modal image segmentation model using split learning.
We use the \textit{Medical Segmentation Decathlon} MSD\footnote{\url{http://medicaldecathlon.com}} brain tumor segmentation dataset (Task 1) to simulate this setup. Each 3D volume in the dataset contains four MRI modalities, namely T1-weighted, post-Gadolinium contrast T1-weighted, T2-weighted, and T2 Fluid-Attenuated Inversion Recovery volumes~\cite{simpson2019large}. For the purpose of this study, we extract one axial slice from each volume through the center of the tumor and formulate the task as a 2D semantic segmentation problem, resulting in a total of 484 images with ground truth annotation masks. We randomly split the data into 338 training, 49 validation, and 97 testing images, corresponding to 70\%, 10\%, 20\% of the data, respectively.
The segmentation task is to predict the brain tumor sub-regions, i.e., edema, enhancing, and non-enhancing tumor. Therefore, our network predicts four output classes, including the background, using a final softmax activation. Given the $K=4$ MRI input modalities, we simulate the Split-U-Net to be trained collaboratively among four sites, as shown in Fig.~\ref{fig:setup}. Each site possesses the images for all patients but for just one modality. Site-1 is assumed to also have the annotation masks and can therefore compute the objective function using a combined Dice loss and cross-entropy loss.
\paragraph{\textbf{Data leakage of shared activation maps:}}
First, we investigate how much data the activation map at each layer can leak when sharing them during Split-U-Net training. We invert all activation maps of a mini-batch from layers $x_0, x_1, x_2, x_3, x_4$, respectively. One can observe that the amount of data leakage reduces with the depth of the network and the resolution of the share activation map. The first level $x_0$ with a resolution similar to the input image is practically non-distinguishable from the original augmented images fed to the encoder networks during training. All inversions computed in this work used $\alpha_{\text{act}}=1e-3$, $\alpha_{\text{tv}}=1e-4$, and $\alpha_{l_2}=1e-5$ (see Eq. \ref{equ:inv_ioss}). We used the Adam optimizer to solve Eq. \ref{equ:inv_img} using a cosine learning rate decay with an initial rate of 0.1\footnote{\textit{\textbf{Implementation:}} We utilize components from MONAI\footnote{\url{https://monai.io/}} and NVIDIA FLARE\footnote{\url{https://developer.nvidia.com/flare}} to implement our SL simulation. In particular, we utilize MONAI's \textit{BasicUNet} as basis for Split-U-Net. All experiments were run on NVIDIA V100 GPUs with 16 GB memory.}.
%
\paragraph{\textbf{Collaborative multi-modal image segmentation:}}
To evaluate the effectiveness of Split-U-Net, we compare it to a baseline U-Net model taking the four MRI modalities directly as input (see the default setup in Table~\ref{tab:network}). In Table~\ref{tab:results}, we show Split-U-Net performs on par with its centralized counterpart (U-Net). The performance is comparable\footnote{The current \href{https://decathlon-10.grand-challenge.org/evaluation/challenge/leaderboard}{leading entry - \texttt{Swin\_UNETR}~\cite{hatamizadeh2022swin}} achieves an average Dice score of 0.647 for the three foreground tumor classes.} to the MSD challenge results reported for 3D tumor segmentation~\cite{antonelli2021medical}.
%
%
\begin{table}[htbp]
\centering
\scriptsize
\caption{Comparison of a centralized U-Net and different Split-U-Net settings with different privacy-preserving measures (dropout and differential privacy (DP)). The setting ``w'' and ``w/o'' indicates the performance of Split U-Net with and without skip connections, respectively; ``$x_3$, $x_4$ only'' indicates the performances when only activations from later layers are being shared. The best Dice score achieved with Split-U-Net for each data subset is highlighted in \underline{\textbf{bold}}.\label{tab:results}}
\begin{tabular}{lrrr}
\toprule
\textbf{Dice}                    & \multicolumn{1}{l}{\textbf{Training (n=338)}} & \multicolumn{1}{l}{\textbf{Validation (n=49)}} & \multicolumn{1}{c}{\textbf{Testing (n=97)}} \\
\hline
U-Net                            & \textit{0.732}                                         & \textit{0.701}                                          & \textit{0.698}                                       \\
Split-U-Net (w/o skip)           & 0.743                                         & \cellcolor[HTML]{FFFFFF}0.619                  & \cellcolor[HTML]{FFFFFF}0.599               \\
Split-U-Net (w skip)             & \underline{\textbf{0.882}}                                         & 0.663                                          & 0.693                                       \\
Split-U-Net ($x_3$, $x_4$ only)        & 0.821                                         & 0.675                                          & 0.650                                       \\
\hline
Split-U-Net ($p_\mathrm{dropout}$=0.1)        & 0.818                                         & 0.648                                          & 0.681                                       \\
Split-U-Net ($p_\mathrm{dropout}$=0.2)        & 0.843                                         & 0.658                                          & 0.683                                       \\
Split-U-Net ($p_\mathrm{dropout}$=0.5)        & 0.766                                         & 0.637                                          & 0.665                                       \\
Split-U-Net ($p_\mathrm{dropout}$=0.8)        & 0.719                                         & 0.643                                          & 0.650                                       \\
\hline
Split-U-Net (DP $\sigma$=1)          & 0.865                                         & 0.671                                          & 0.691                                       \\
Split-U-Net (DP $\sigma$=2)          & 0.797                                         & 0.669                                          & \underline{\textbf{0.695}}                              \\
Split-U-Net (DP $\sigma$=3)          & 0.821                                         & 0.658                                          & 0.666                                       \\
Split-U-Net (DP $\sigma$=5)          & 0.811                                         & \underline{\textbf{0.684}}                                          & 0.687                    \\                  
Split-U-Net (DP $\sigma$=50)         & 0.543                                         & 0.394                                                &	0.393 \\
\bottomrule
\end{tabular}
\end{table}
%
\paragraph{\textbf{Effectiveness of defenses:}}
Adding dropout and Gaussian noise during training can be an effective defense (Fig.~\ref{fig:defense_inv}). The SSIM scores between originals and inversions go down with higher $p_\mathrm{dropout}$ or $\sigma$ as shown in Fig.~\ref{fig:ssim}. The model performance is less affected when adding DP and even benefits from it during training, as seen in Table~\ref{tab:network} for $\sigma=2.0$ in contrast to using dropout as a defense.
%

\section{Discussion}
\label{sec:discussion}
\noindent To the best of our knowledge, our work was the first to apply SL to a multi-modal image segmentation task. We showed competitive results of Split-U-Net for 2D brain tumor segmentation on a relatively small dataset (only one slice per original volume). Further hyperparameter tuning and data augmentation might improve the performance. It should be investigated if weight sharing between the encoder branches could allow for further performance boosts~\cite{thapa2020splitfed}. An extension of Split-U-Net to 3D semantic segmentation tasks would be straightforward.

A major focus of this work is on the security aspect when applying SL. As shown in our results, depending on the depth of activation layers inside Split-U-Net, the data inversion attack can be successful, generating inversions that are visually indistinguishable from the original images (SSIM close to 1.0). This is the case, especially for the first layer ($x_0$). Finding an appropriate defense strategy against such inversion attacks is very important.
It can be assumed that the same defense settings are effective for each modality used in Split-U-Net training. Therefore, a recommendation would be for the site possessing both images and labels to study the data leakage vulnerabilities using our proposed data inversion and data leakage metrics to establish a secure setting that each collaborator can use. Of course, this assumes a level of trust in this site but might help protect against a potentially malicious server that coordinates the split learning. At the same time, each site could utilize public datasets with images and labels, as we have done in this study, to measure the data leakage risks of the network architecture they would like to train in real-world SL.
Our results indicate that the dangers come from the architecture itself rather than the particular dataset used for training (see Fig.~\ref{fig:skip_invs} where the inversion quality is not affected by different samples in the batch). This is in contrast to other studies in horizontal FL, where certain images in the batch are more likely to leak data~\cite{yin2021see,hatamizadeh2022gradient}. 

Our study also has some limitations. For example, the inversion attack assumes to have access to the current state of the model that the data site uses to compute its forward pass ($\fenc^k(X)$). This setting is typically referred to as a ``white-box'' attack~\cite{he2019model} and assumes the attacker has knowledge about the state of the model during training. This could be true in some implementations of SL where one of the participants sends an initialization for all participants. As the training continues, this initial model will become less and less useful to the attacker. At the same time, our finding shows that a potential avenue for more secure implementations of SL is to not use a common initialization but let each participant randomly initialize their part of the model. A ``black-box'' attack~\cite{he2019model} where the inversion needs to optimize for both the inputs and the current state of the model could be implemented next to better measure the data leakage risks in such a scenario.
Furthermore, we assumed the participating sites to have a common anonymous identifier used to build mini-batches with images of corresponding patients. In real-world scenarios, a pre-processing step to securely compute the intersecting set of patients between sites has to be performed~\cite{angelou2020asymmetric}. Also, some synchronization of data augmentation across different modalities should be incorporated in the communication protocols.
An additional privacy risk in SL is the inversion of label sets from the shared model gradients. A similar attack to the one presented in this work could be applied to match gradients during SL to recover the label masks. However, we assumed that tumor segmentation masks are less likely to leak patient-identifiable information and therefore focused on the data/image recovery in this work.
In this work, we simulated a multi-site FL study using pre-registered multi-modal MRI scans. In reality, more variations that are potentially critical to model performance would need to be considered before performing similar collaborative model training, including temporal and spatial misalignment across images of the same patient and mismatch between image and annotation masks.
Finally, cryptographic techniques like homomorphic encryption~\cite{zhang2020batchcrypt} or secure multi-party computation~\cite{kaissis2020secure} could be employed to reduce the risk of data leakage in SL. Those techniques typically come with higher computation costs but should be explored, especially for medical image analysis tasks where patient privacy is of utmost concern.

In conclusion, we provided strong evidence that SL can be useful for biomedical image segmentation tasks when taking the appropriate security considerations into account. A real-world implementation of SL will provide clinical collaborators the chance to jointly leverage all available data to train more robust and generalizable AI models.

\clearpage
\newpage
\bibliographystyle{splncs04}
\bibliography{refs}

\end{document}